\documentclass{clv2-eng}

\usepackage{url}
\usepackage{latexsym}
\usepackage{multirow}
\usepackage{comment}

\usepackage{xcolor}
\usepackage{booktabs}
\usepackage{multirow}
\usepackage{graphicx}

\newcommand\mf[1]{\textcolor{black}{#1}}
\newcommand\sml[1]{\textcolor{black}{#1}}
\newcommand\todo[1]{\textcolor{black}{#1}}

\issue{1}{1}{2018}
\title{Multilingual Neural Machine Translation for Zero-Resource Languages}

\runningtitle{Multilingual NMT for Zero-Resource Languages}

\runningauthor{Lakew et al.}

\begin{document}

\author{Surafel M. Lakew\thanks{Fondazione Bruno Kessler, Via Somarive 18, 38123 Povo (Trento), Italy. E-mail:~\texttt{lakew@fbk.eu}}}
\affil{Fondazione Bruno Kessler\\University of Trento}

\author{Marcello Federico}
\affil{MMT Srl, Trento\\Fondazione Bruno Kessler}

\author{Matteo Negri}
\affil{Fondazione Bruno Kessler}

\author{Marco Turchi}
\affil{Fondazione Bruno Kessler}
\maketitle

\begin{abstract}
In recent years, Neural Machine Translation (NMT) has been shown to be more effective than phrase-based statistical methods, thus quickly becoming the state of the art in machine translation (MT). 
However, NMT systems are limited in translating low-resourced languages, due to the significant amount of parallel data that is required to learn 
useful mappings between languages. 
In this work, we show how the so-called multilingual NMT can help to tackle the challenges associated with low-resourced language translation. The underlying principle 
of multilingual NMT is to force the creation of hidden representations of words in a shared semantic space across multiple languages, thus enabling a positive parameter transfer across languages. 
Along this direction, we present multilingual translation experiments with three languages (English, Italian, Romanian) covering six translation directions, utilizing both recurrent neural networks and transformer (or self-attentive) neural networks. We then 
focus on the zero-shot translation problem, that is how to leverage multi-lingual data in order to learn translation directions 
that are not covered by the available training material. 
To this aim,
we introduce our recently proposed iterative self-training  method, which incrementally improves a multilingual NMT 
on a zero-shot direction by just relying on monolingual data. Our results on TED talks data show that multilingual NMT outperforms conventional bilingual NMT, that 
the transformer NMT outperforms recurrent NMT, and that zero-shot NMT   
outperforms conventional pivoting methods and even matches the performance of a fully-trained bilingual system. 
\end{abstract}

\section{Introduction}
Neural machine translation (NMT) has shown its effectiveness by delivering the best performance in the IWSLT~\cite{cettolo2016iwslt} and WMT~\cite{bojar2016findings} evaluation campaigns of the last three years. Unlike rule-based or statistical MT, the end-to-end learning approach 
of NMT models the mapping from source to target language directly through a posterior probability. 
The essential component of an NMT system includes an encoder, a decoder and an attention mechanism~\cite{bahdanau2014neural}. Despite the continuous improvement in performance and translation quality~\cite{bentivogli:CSL2018}, NMT models are highly dependent on the availability of extensive parallel data, which in practice can only be acquired for a very limited number of language pairs. 
For this reason,  building effective NMT systems for low-resource languages becomes a primary challenge~\cite{koehn2017six}. In fact,~\namecite{zoph2016transfer} showed how a standard string-to-tree statistical MT system~\cite{galley2006scalable} can effectively outperform NMT methods for low-resource languages, such as Hausa, Uzbek, and Urdu. 

\begin{figure}[!t]\centerline{\epsfig{figure=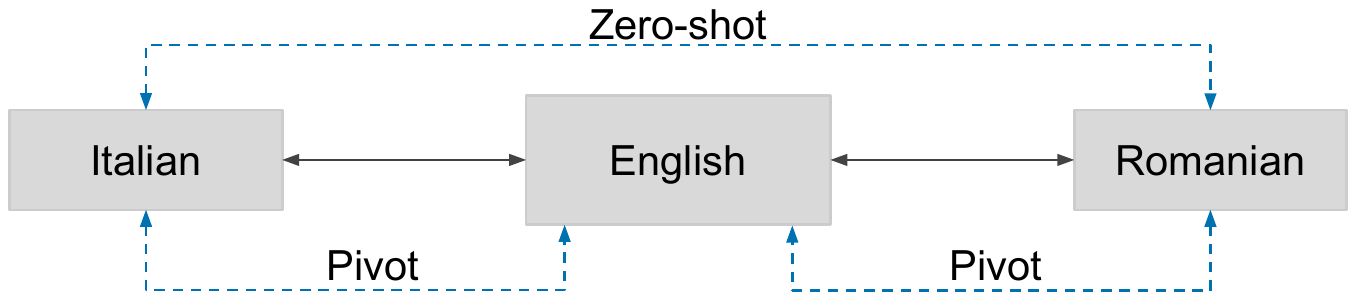,width=4in}}
\caption{{\it A multilingual setting with parallel training data in four translation directions: Italian$\leftrightarrow$English and Romanian$\leftrightarrow$English. Translations between Italian and Romanian are either inferred directly (zero-shot) or by translating through English (pivoting).}}
\label{figure:scenario}
\end{figure}

\noindent
In this work, we approach low-resource machine translation with so-called {\em multilingual} NMT~\cite{johnson2016google,ha2016toward}, which considers the 
use of NMT to target many-to-many translation directions. Our motivation is that 
intensive cross-lingual transfer~\cite{Terence:LT} via parameter sharing 
across multiple languages should ideally help in the case of similar languages and sparse training data. 
In particular, we investigate multilingual NMT across Italian, Romanian, and  English, and simulate low-resource conditions by limiting the amount of available parallel data.

\noindent
Among the various approaches for multilingual NMT, the simplest and most effective one is to train a single neural network on parallel data including multiple translation directions and to prepend to each source sentence a \textit{flag}  specifying the desired target language~\cite{ha2016toward,johnson2016google}. 
In this work, we investigate multi-lingual NMT under low-resource conditions and with two popular NMT architectures: recurrent LSTM-based NMT and the self-attentive or transformer NMT model. 
In particular, we train and evaluate our systems on a collection of TED talks~\cite{cettoloEtAl:EAMT2012}, over six translation directions: English$\leftrightarrow$Italian, English$\leftrightarrow$Romanian, and Italian$\leftrightarrow$Romanian.

A major advantage of 
multi-lingual NMT is the possibility to perform a zero-shot translation, that is to query the system on a direction for which no training data was provided. The case we consider is illustrated in Figure~\ref{figure:scenario}: we assume to only have Italian-English and English-Romanian training data and that we need to translate between Italian and Romanian in both directions. To solve this task, we propose a {\em self-learning} method that permits a multi-lingual NMT system trained on the above mentioned language pairs to progressively learn how to translate between Romanian and Italian directly from its own translations. 
We show that our zero-shot self-training approach not only improves over a conventional pivoting approach, by bridging Romanian-Italian through English (see Figure~\ref{figure:scenario}), but that it also matches the performance of bilingual systems trained on Italian-Romanian data. 
The contribution of this work is twofold:\footnote{This paper integrates and extends work presented in \cite{MNMTlow-resourceSurafel} and \cite{lakew2017improving}.}
\begin{itemize}
\item Comparing RNN and Transformer approaches in a multilingual NMT setting;
\item A self-learning approach to improve the zero-shot translation task of a multilingual model.
\end{itemize}

The paper is organized as follows. In Section \ref{related}, we present previous works on multilingual NMT, zero-shot NMT, and NMT training with self-generated data. In Section \ref{sec:NMT}, we  introduce the two prominent NMT approaches evaluated in this paper, the recurrent and the transformer models.  In Section \ref{sec:0shot}, we introduce our 
multilingual NMT approach and our self-training method for zero-shot learning. In Section \ref{sec:exp}, we describe our experimental set-up and the NMT model configurations. In Section \ref{results}, we present and discuss the results of our experiments. Section \ref{sec:concl} ends the paper with our conclusions.

\section{Previous Work}\label{related}
\subsection{Multilingual NMT}

Previous works in multilingual NMT are characterized by the use of separate encoding and/or decoding networks for each translation direction. \namecite{dong2015multi} proposed  a multi-task learning approach for a 
\textit{one-to-many} translation scenario, by sharing hidden representations 
among related tasks -- \textit{i.e} the source languages -- to enhance 
generalization on the target language. In particular, 
they used a single encoder for all source languages and separate attention mechanisms and decoders for every target language. In a related work,~\namecite{luong2015multi}
used distinct encoder and decoder networks for modeling language pairs in a 
\textit{many-to-many} setting. 
Aimed at reducing ambiguities at translation time, \namecite{zoph2016multi} employed a \textit{many-to-one}  
system that considers two languages on the encoder side and one target language on the decoder side. In particular, the attention model is applied to a combination of the two encoder states.
In a \textit{many-to-many} translation scenario, \namecite{firat2016multi} introduced a way to share the attention mechanism across multiple languages. As in \namecite{dong2015multi}  (but 
only on the decoder side) and in \namecite{luong2015multi}, they used separate encoders and decoders for each source and target language.

Despite the reported improvements, the need of using an additional encoder and/or decoder for every language added to the system tells the limitation of these approaches, by making their network complex and expensive to train.

In a very different way, \namecite{johnson2016google} and \namecite{ha2016toward} developed 
similar multilingual NMT approaches by introducing a 
\textit{target-forcing} token in the input. The approach in \namecite{ha2016toward} applies a language-specific code to words from different languages in a mixed-language vocabulary. 
In practice, they force the decoder to translate into a specific target language by prepending and appending  an artificial token to the source text.  
However, their word and sub-word level language-specific coding mechanism 
significantly increases the input length, which shows to have an impact 
on the computational cost and performance of NMT~\cite{cho2014properties}. 
In \namecite{johnson2016google}, only one artificial token is pre-pended to the entire source sentences in order to specify the target language. Hence, the same token is also used to trigger the decoder generating the translation (cf. Figure~\ref{figure:enc-dec}). 
Remarkably, pre-pending language tokens to the input string has greatly simplified multi-lingual NMT, 
by eliminating the need of having separate encoder/decoder networks and attention mechanism for every 
new language pair.

\subsection{Zero-Shot and Self-Learning } 
Zero-resource NMT has been proposed in \cite{firat2016zero} and it extends the work by \namecite{firat2016multi}. The authors proposed a \textit{many-to-one} translation setting and used the idea of generating a pseudo-parallel corpus~\cite{sennrich2015improvingMono}, using a pivot language, to fine tune their model. However, also in this case the 
need of separate encoders and decoders for every language pair 
significantly increases the complexity of the model.

An attractive feature of the \textit{target-forcing} mechanism  comes from the possibility to perform zero-shot translation with the same multilingual setting as in \cite{johnson2016google,ha2016toward}.
Both the works reported that a multilingual system trained on a large amount of data improves over a baseline bilingual model and that it is also capable of performing zero-shot translation, assuming that the zero-shot source and target languages have been observed during training paired with some other languages.

However, recent experiments have shown that the mechanism fails to achieve reasonable zero-shot translation performance for low-resource languages~\cite{MNMTlow-resourceSurafel}. The promising results in \cite{johnson2016google} and \cite{ha2016toward} hence require further investigation to verify if their method can work in various language settings, particularly across distant languages. 

An alternative approach to zero-shot translation in a resource-scarce scenario is to use a pivot language \cite{cettolo2011bootstrapping}, that is, using an intermediate language for translation. 
While this solution is usually pursued by deploying two or more bilingual models, in this work we aim to achieve comparable results using a single multilingual model.

Training procedures using synthetic data have been around for a while. For instance, in statistical machine translation (SMT), \namecite{oflazer2007exploringIncremental} and \namecite{bechara2011statisticalIncremental} showed how the output of a translation model can be used iteratively to improve results in a task like post-editing. Mechanisms like back-translating the target side of a single language pair have been used for domain adaptation~\cite{bertoldi2009domain} and more recently by \namecite{sennrich2015improvingMono}  to improve an NMT baseline model. In \cite{dual-learningMT}, a dual-learning mechanism is proposed where two NMT models working in the opposite directions provide each other feedback signals
that permit them to learn from monolingual data. In a related way, our approach 
also considers training from monolingual data.
As a difference, however, our proposed method
leverages the capability of the network to jointly learn multiple translation directions and to directly generate the translations used for self-training.

Although our brief survey shows that re-using the output of an MT system for further training and improvement has been successfully applied in different settings, 
our approach differs from past works in  two aspects: 
\textit{i)}  introducing a new self-training method integrated in a multilingual NMT architecture, and \textit{ii)} casting the approach into a \textit{self-correcting}  procedure over two dual zero-shot directions, so that incrementally improved translations mutually reinforce each direction.

\section{Neural Machine Translation}
\label{sec:NMT}
\begin{figure}\centerline{\epsfig{figure=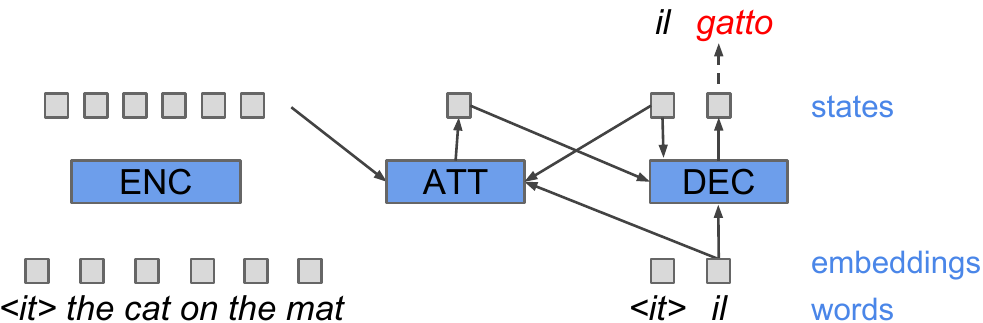,width=4in}}
\caption{{\it Encoder-decoder-attention NMT architecture. Once the encoder has generated his states for all the input words, the decoder starts  generating the translation word by word. The target word "gatto" (cat) is generated from the previously generated target word "il" (the), the previous decoder state and the context state. The context state is a selection and combination of encoder states computed by the attention model. Finally, notice that the target-language forcing symbol "<it>" (Italian), prepended to the input, is also used to trigger the first output word.}}
\label{figure:enc-dec}
\end{figure}

State-of-the-art NMT systems comprise an encoder, a decoder, and an attention mechanism, which are jointly trained with maximum likelihood in an end-to-end fashion~\cite{bahdanau2014neural}. Among the different variants, two popular ones
are the recurrent NMT~\cite{sutskever2014sequence} and the transformer NMT~\cite{vaswani2017attention} models. In both the approaches, the encoder is purposed to map a source sentence into a sequence of state vectors, whereas the decoder uses the previous decoder states, its last output, and the attention model state  to infer the next target word (see Figure~\ref{figure:enc-dec}). In a broad sense, the attention mechanism selects and combines the encoder states that are
most relevant to infer the next word~\cite{luong2015effective}. In our multi-lingual setting, the decoding process is triggered by specifying the target language identifier (Italian, {\em <it>}, in the example of Figure~\ref{figure:enc-dec}).
In the following two sub-sections, we briefly summarize the main features of the two considered architectures.

\subsection{Recurrent NMT}\label{subsec:nmt_rnn}
Recurrent NMT models employ recurrent neural networks (RNNs) to build the internal representations of both the 
encoder and decoder. Recurrent layers are in general implemented with LSTM~\cite{hochreiter1997long} or GRU \cite{cho2014properties} units, which include gates able to control the propagation of information over time. 
While the encoder typically uses a bi-directional RNN, so that both left-to-right and right-to-left word dependencies 
are captured (see left-hand of Figure~\ref{figure:rnn-transformer}), the decoder by design can only learn left-to-right dependencies. In general, deep recurrent NMT is achieved by stacking multiple recurrent 
layers inside both the encoder and the decoder.

While RNNs are in theory the most expressive type of neural networks~\cite{siegelmann1995computational}, they are in practice hard and slow to train. In particular, the combination of two levels of deepness, horizontal along time 
and vertical across the layers, makes gradient-based optimization of deep RNNs particularly slow to converge and difficult to parallelize \cite{wu2016google}. Recent work succeeded in speeding up training convergence \cite{sennrich-EtAl:2017:WMT} of recurrent NMT by reducing the network size via parameter tying and layer normalization. 
On the other hand, the {\em simple recurrent} NMT model proposed by \cite{DiGangi-Federico:2018:EAMT}, which weakens the network time dependencies, 
has shown to outperform LSTM-based NMT both in training speed and performance.

\subsection{Transformer NMT}\label{subsec:nmt_tranformer}
\begin{figure}\centerline{\epsfig{figure=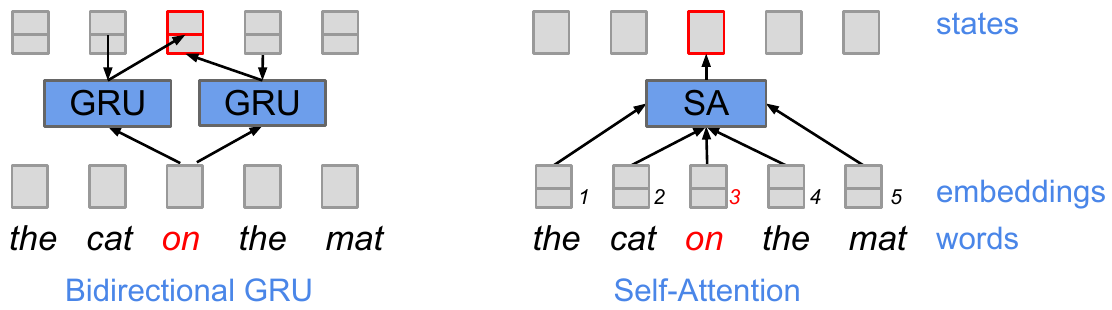,width=5in}}
\caption{{\it Single-layer encoders with recurrent (left) and transformer networks (right). A bi-directional 
recurrent encoder generates the state for word "on" with two GRU units. Notice that states must be generated 
sequentially. The transformer generates the state of word "on" with a self-attention model that looks at 
all the input embeddings, which are extended with position information. Notice that all the states can be generated independently.}}
\label{figure:rnn-transformer}
\end{figure}

The transformer architecture \cite{vaswani2017attention} works by relying on a self-attention 
mechanism, removing all the recurrent operations that are found in the RNN case~\cite{vaswani2017attention}. In other words, the attention mechanism is re-purposed to also compute the latent space representation of both the encoder and the decoder. The right-hand side of Figure~\ref{figure:rnn-transformer} depicts a simple one-layer encoder based on self-attention. 
Notice that, in absence of recurrence, a \textit{positional-encoding} is added to the input and output embeddings. Similarly, as the time-step in RNN, the positional information provides the transformer network with the order of input and output sequences. In our work we use 
absolute positional encoding but, very recently, the use of 
relative positional information has been shown to improve the network performance~\cite{shaw2018self}.

Overall, the transformer is organized as a stack of encoder-decoder networks that works in an auto-regressive way, using the previously generated symbol as input for the next prediction. Both the decoder and encoder can be composed of uniform layers, each built of sub-layers, i.e., a multi-head self-attention sub-layer and a position-wise feed-forward network sub-layer. Specifically for the decoder, an extra multi-head attentional 
layer is added to attend to the output states of the encoder.
Multi-head attention layers enable the use of multiple attention functions with a computational cost similar to utilizing a single attention. 

\noindent

\section{Zero-Shot Self-Training in Multilingual NMT}
\label{sec:0shot}

\begin{figure*}\centerline{\epsfig{figure=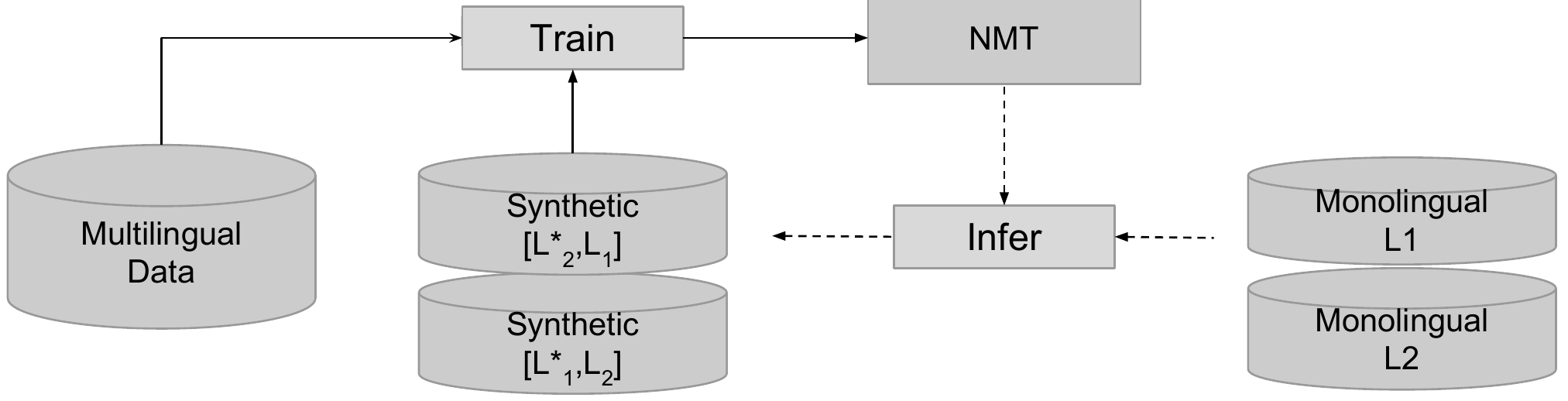,width=5.3in}}
\caption{{\it Illustration of the proposed multilingual \emph{train-infer-train} strategy. Using a standard NMT architecture, a portion of two zero-shot directions monolingual dataset is extracted for inference to construct a dual source$\leftrightarrow$target mixed-input and continue the training. The solid lines show the training process, whereas the dashed lines indicate the inference stage.}}
\label{fig:train-infer-train_illustration}
\end{figure*}

In this setting, our goal is to improve translation in the zero-shot directions of a baseline multilingual model  trained on data covering \textit{n} languages but not all their possible combinations (see  Figure~\ref{figure:scenario}). 
\todo{After training a baseline multilingual model with the target-forcing method \cite{johnson2016google}, our self-learning approach works in the following way: 
\begin{itemize}
\item First, a dual zero-shot inference (i.e., source$\leftrightarrow$target directions) is performed utilizing monolingual data extracted from the training corpus;
\item Second, the training resumes combining the inference output and the original multilingual data from the non zero-shot directions;
\item Third, this cycle of \textit{training-inference-training} is repeated until a convergence point is reached on the dual zero-shot directions.
\end{itemize}
Notice that, at each iteration, the original training data is augmented only with the last batch of generated translations. We observe that the generated outputs initially contain a mix of words from the shared vocabulary but, after few iterations, they tend to only contain words in the zero-shot target language thus becoming more and more suitable for learning.}
The training and inference strategy of the proposed approach is summarized in Algorithm~\ref{table:training_strategy}, whereas the flow chart (see Figure~\ref{fig:train-infer-train_illustration}) further illustrates the training and inference pipeline.

\begin{table}[h!]\caption{\label{table:training_strategy} {\it Self-training algorithm for zero-shot directions $l_1\leftrightarrow l_2$.}}
\vspace{2mm}
\centering
\begin{tabular}{l}
\textbf{Algorithm 1:} Train-Infer-Train (TIF)\\ 
\toprule
1: TIF: ($D$, $l_1$, $l_2$) \\
2: \hspace{2mm} $M$ $\leftarrow$ Train($\emptyset$, $D$)\hspace{3cm} //train multilingual base model on data $D$ \\
3: \hspace{2mm} $L_1$ $\leftarrow$ Extract($D$, $l_1$)\hspace{2.5cm} //extract $l_1$ monolingual data from $D$\\
4: \hspace{2mm} $L_2$ $\leftarrow$ Extract($D$, $l_2$)\hspace{2.5cm} //extract $l_2$ monolingual data from $D$\\
5: \hspace{2mm} \textbf{for} $i$ $=$ $1$, $N$ \textbf{do} \\
6: \hspace{5mm} $L_2^*$ $\leftarrow$ Infer($M$, $L_1$, $l_2$)\hspace{2cm} //translate $L_1$ into $l_2$ \\ 
7: \hspace{5mm} $L_1^*$ $\leftarrow$ Infer($M^*$, $L_2$, $l_1$)\hspace{2cm} //translate $L_2$ into $l_1$ \\ 
8: \hspace{5mm} $D^*$ $\leftarrow$ $D \cup (L_1^*,L_2) \cup (L_2^*,L_1)$\hspace{1cm} //augment original data\\
9:\hspace{5mm} $M$ $\leftarrow$ Train($M$, $D^*$)\hspace{2.5cm} //re-train model on augmented data \\
10: \hspace{0.5mm} \textbf{end for} \\
11: \hspace{0.5mm} return $M$  \\ 
\end{tabular}
\end{table}

\noindent
The proposed approach is performed in three steps, where the latter two are iterated for a few rounds. 
In the first step \sml{(line 2)}, a multilingual NMT system $M$ is trained from scratch on the available data $D$ (''\textit{Train}'' step).
In the second step \sml{(lines 7-8)}, the last trained model $M$ 
is run to translate (''\textit{Infer}'' step) between the zero-shot directions monolingual data $L_1$ and $L_2$ extracted from
$D$ \sml{(lines 3-4)}. Then, in the third step \sml{(line 10)}, training of $M$ is re-started on the original 
data $D$ plus the generated synthetic translations $L_2^*$ and $L_1^*$, 
by keeping the extracted monolingual data $L_1$ and $L_2$ always on the target side (''\textit{Train}'' step). The updated model
is then again used to generate synthetic translations, on which to re-train $M$, and so on.

In the multilingual NMT scenario, the automatic translations used as the source part of the extended training data will likely contain a mixed-language that includes words from a vocabulary shared with other languages. The expectation is that, round after round, the model will generate better outputs by learning at the same time to translate and ``correct'' its own translations by removing spurious elements from other languages. If this intuition holds, 
the iterative improvement will yield increasingly better results in  translating between the source $\leftrightarrow$ target zero-shot directions.

\section{Experiments}
\label{sec:exp}
\subsection{NMT Settings}

We trained and evaluated multilingual NMT systems based on the RNN~\cite{cho2014learningGRU} and  transformer~\cite{vaswani2017attention} models. Table~\ref{table:parameters} summarizes the hyper-parameters used for all our models. The RNN experiments are carried out using the NMT toolkit Nematus\footnote{https://github.com/EdinburghNLP/nematus}~\cite{sennrich2017nematus}, whereas the transformer models are trained using the open source OpenNMT-tf\footnote{https://github.com/OpenNMT/OpenNMT-tf} toolkit~\cite{klein2017opennmt}. 

Training and inference hyper-parameters for both approaches and toolkits are fixed as follows.
For the RNN experiments, the Adagrad~\cite{duchi2011adaptive} optimization algorithm is utilized with an initial learning rate of 0.01 and mini-batches of size 100. Considering the high data sparsity of our low-resource setting and to prevent model over-fitting~\cite{srivastava2014dropout}, we applied a dropout on every layer, with probability $0.2$ on the embeddings and the hidden layers, and $0.1$ on the input and output layers. For the experiments using the transformer approach, a dropout of $0.3$ is used globally. To train the baseline multilingual NMT, we use Adam~\cite{kingma2014adam} as the optimization algorithm with an initial learning rate scale constant of $2$. For the transformer, the learning rate is increased linearly in the early stages (\emph{warmup\_training\_steps=$16,000$}); after that, it is decreased with an inverse square root of training step~\cite{vaswani2017attention}. 
In all the reported experiments, the baseline models are trained until convergence, while each training round after the inference stage is assumed to iterate for 3 epochs. In case of the transformer NMT, M4-NMT (four translation directions multilingual system), and M6-NMT (six translation directions multilingual system)    
BLEU scores are computed using averaged model from the last seven checkpoints in the 
same training run~\cite{junczys2016amu}. For decoding, a beam search of size 10 is applied for recurrent models, whereas one of size $4$ is used for transformer models.

\begin{table}[h!]\caption{\label{table:parameters} {\it Hyper-parameters used to train RNN and transformer models, unless differently specified.}}
\centerline{
\begin{tabular}{l c c c c c r}
\hline
\multicolumn{1}{c}{}  & enc/dec & embedding & hidden & encoder & decoder & batch \\
\multicolumn{1}{c}{}  & type    & size      & units  & depth   & depth   & size \\ \hline
			RNN & GRU & 1024 & 1024 & 2 & 2 & 128 seg \\
			Transformer & Transformer & 512 & 512 & 6 & 6 & 4096 tok \\ \hline
\end{tabular}}
\end{table}

\subsection{Dataset and preprocessing}
We run all our experiments on the multilingual translation shared task data released for
the 2017 International Workshop on Spoken Language Translation (IWSLT)\footnote{http://workshop2017.iwslt.org/}. 
In particular, we used the subset of training data covering all possible language pair combinations between Italian, 
Romanian, and English~\cite{cettoloEtAl:EAMT2012}. For development and evaluation of the models, we used the corresponding sets from the IWSLT2010~\cite{paulIwslt2010overview} and IWSLT2017 evaluation campaigns. Details about the used data sets are reported in Table~\ref{data-set}.
At the preprocessing stage, we applied word segmentation for each training condition (i.e. bilingual or multi-lingual)
by learning a sub-word dictionary via Byte-Pair Encoding~\cite{sennrich2015neuralSub-word}, by setting the number of merging rules to 39,500. We observe a rather high overlap between the language pairs (i.e the English dataset paired with Romanian is highly similar to the English paired with Italian). Because of this overlapping, the actual unique sentences in the dataset are approximately half of the total size. Consequently, on one side, this exacerbates the low-resource aspect in the multilingual models while, on the other side, we expect some positive effect on the zero-shot condition. The final size of the vocabulary, both in case of the bilingual and the multilingual models, stays under 40,000 sub-words. An evaluation script to compute the BLEU~\cite{papineni2002bleu} score is used to validate models on the dev set and 
later to choose the best performing models. Furthermore, significance tests computed for the BLEU scores are reported using Multeval~\cite{clark2011better}.

\begin{table}[h!]
\caption{\label{data-set} The total number of parallel sentences used for training, development, and test in our low-resource scenario.}
\begin{center}
\begin{tabular}{l c  c r}
\hline \it Language Pair & \it Train & \it Test10 & \it Test17 \\ \hline 
En-It & 231619  & 929 & 1147 \\
En-Ro & 220538  & 929 & 1129 \\ 
It-Ro & 217551  & 914 & 1127 \\ \hline
\end{tabular}
\end{center}
\end{table}

\noindent
We trained models for two different scenarios. The first one is the multi-lingual scenario with all the available language pairs, while the second one is for the zero-shot and pivoting approaches which excludes $Romanian-Italian$ parallel sentences
from the training data. For both scenarios, we have also trained bilingual RNN and Transformer
models for comparing bilingual against multilingual systems and for comparing pivoting with bilingual and multilingual models.

\section{Models and Results}
\label{results}
\subsection{Bilingual Vs. Multilingual NMT}
We compare the translation performance of six independently-trained bilingual models against one single multilingual 
model trained on the concatenation of all the six language pairs datasets, after prepending the language flag on the source side of each sentence. The performance of both types of systems is evaluated on test2017 and reported in 
Table~\ref{biVSm-nmt2}.
The experiments show that a multilingual system 
outperforms the bilingual systems with variable margins. The improvements, which are  observed in all the language directions, are likely brought by the cross-lingual parameter transfer between the additional language pairs involved in the source and target side.

\begin{table}[h!]
\centering
\caption{\label{biVSm-nmt2} Comparison between six bilingual models (NMT) against a single multilingual model (M6-NMT) on Test17.}
\begin{tabular}{l c c c c c r}
\hline
\multicolumn{1}{c}{\multirow{2}{*}{\it Direction}} & \multicolumn{3}{c}{\it RNN} & \multicolumn{3}{c}{\it Transformer} \\ \cmidrule(l){2-4}  \cmidrule(l){5-7} 
\multicolumn{1}{c}{}                               &  NMT  & M6-NMT  & $\Delta$  & NMT        & M6-NMT     & $\Delta$     \\ \hline
En $\rightarrow$ It                                & 27.44    & 28.22  & +0.78   & 29.24      & 30.88     & +1.64      \\
It $\rightarrow$ En                                & 29.9     & 31.84  & +1.94   & 33.12      & 36.05     & +2.93      \\
En $\rightarrow$ Ro                                & 20.96    & 21.56  & +0.60   & 23.05      & 24.65     & +1.60      \\
Ro $\rightarrow$ En                                & 25.44    & 27.24  & +1.80   & 28.40      & 30.25     & +1.85      \\
It $\rightarrow$ Ro                                & 17.7     & 18.95  & +1.25   & 20.10      & 20.13     & +0.03      \\
Ro $\rightarrow$ It                                & 19.99    & 20.72  & +0.73   & 21.36      & 21.81     & +0.45      \\ \hline
\end{tabular}
\label{table:bi-multi}.
\end{table}

\noindent
Table~\ref{biVSm-nmt2} shows that the transformer model is definitely superior to the RNN model for all directions
and set-ups. With a larger margin of +3.22 (NMT) and +4.21 (M6-NMT), the transformer outperforms the RNN in the It-En direction. The closest performance between the two approaches is observed in the Ro-It direction, with the transformer showing a +1.37 (NMT) and +1.09 (M6-NMT) 
BLEU score increase
compared to the RNN counterpart.  
Moreover, multilingual architectures in general outperform their equivalent models trained on single language pairs. The highest improvement of the M6-NMT over the NMT systems is observed when the target language is English. For instance, in the It-En direction, the multilingual approach gained +1.94 (RNN) and +2.93 (Transformer) over the single language pair models. Similarly, a +1.80 (RNN) and +1.85 (Transformer) gains are observed in the Ro-En direction. However, the smallest gain of the multilingual models occurred when translating into either Italian or Romanian. 
Independently from the target language 
in the experimental setting, the slight difference in the dataset size (that tends to benefit the English target, see Table \ref{data-set}) showed to impact the performance on non-English target directions.

\subsection{Pivoting using a Multilingual Model}
The pivoting experiment is setup by dropping the Italian$\leftrightarrow$Romanian parallel segments from the training data, and by training \textit{i)} a new multilingual-model covering four directions and \textit{ii)} a single model for each language direction (It $\rightarrow$ En, En $\rightarrow$ It, Ro $\rightarrow$ En, En $\rightarrow$ Ro). Our main aim is to analyze how a multilingual model can improve a zero-shot translation task using a pivoting mechanism with English as a bridge language in the experiment. Moreover, the use of a multilingual model for pivoting is motivated by the results we acquired using the M6-NMT (see Table~\ref{biVSm-nmt2}). 

\begin{table}[h!]
\caption{Comparison of pivoting with bilingual models (NMT) and with multilingual models (M4-NMT) on Test17.}
\label{pm-nmt2}
\centering
\begin{tabular}{l c c c c c r}
\hline
\multirow{2}{*}{\textit{Direction}} & \multicolumn{3}{c}{\textit{RNN}}                                                      & \multicolumn{3}{c}{\textit{Transformer}}                                              \\  \cmidrule(l){2-4}  \cmidrule(l){5-7}
                                    & \multicolumn{1}{l}{NMT} & \multicolumn{1}{l}{M4-NMT} & \multicolumn{1}{c}{$\Delta_{\tt NMT}^{\tt M4-NMT}$} & \multicolumn{1}{l}{NMT} & \multicolumn{1}{l}{M4-NMT} & \multicolumn{1}{c}{$\Delta_{\tt NMT}^{\tt M4-NMT}$} \\ \hline
It $\rightarrow$ En	$\rightarrow$ Ro	& 16.3                  & 17.58			& +1.28      &  16.59                      & 16.77               	 & +0.18                           \\
Ro $\rightarrow$ En $\rightarrow$ It        & 18.69                 & 18.66          & -0.03                      &   17.87                                 			&  19.39         & +1.52                            \\ \hline
\end{tabular}
\end{table}

\noindent
The results in Table~\ref{pm-nmt2} show the potential, although partial, of using multilingual models for pivoting unseen translation directions. The comparable results achieved in both directions speak 
in favor of training and deploying one system instead of two distinct NMT systems. \mf{Remarkably, the marked difference between RNN and transformer is vanished in this condition.} Pivoting using the M4-NMT system showed to perform better in three out of four evaluations, from the RNN and transformer runs.  
Note that the performance of the final translation (i.e pivot-target) is subject to the noise that has been propagated from the source-pivot translation step. Meaning pivoting is a favorable strategy when we have strong models in both directions of the pivot language.

\subsection{Zero-shot Translations}
For the direct zero-shot experiments and the application of the \textit{train-infer-train} strategy, we only 
carried out experiments with the transformer approach. Preliminary results showed its superiority over the RNN
together with the possibility to carry out experiments faster and with multiple GPUs. 

\noindent
In this experiment, we show how our approach helps to significantly boost the baseline multilingual NMT model. 
We run the train-infer-train for five consecutive stages, where each round consists in 2-3 epochs of additional training on the augmented training data. Table~\ref{4direction_zst} shows the improvements on the dual Italian$\leftrightarrow$Romanian zero-shot directions.

\begin{table}[h!]
\centering
\caption{\label{4direction_zst} Comparison between a baseline multilingual model (M4-NMT) against the results from our proposed \textit{train-infer-train} approach in a subsequent five rounds for the Italian$\leftrightarrow$Romanian zero-shot directions.}
\begin{tabular}{l c c c c c c}
\hline
\textit{Direction}   & \textit{M4-NMT} & \textit{R1} & \textit{R2} & \textit{R3} & \textit{R4} & \textit{R5} \\ \hline
It$\rightarrow$Ro & 4.72       & 15.22              & 18.46              & 19.31              & 19.59              &  \bf{20.11}            \\
Ro$\rightarrow$It & 5.09       & 16.31              & 20.31              & 21.44              & 21.63              &    \bf{22.41}    		\\ \hline     
\end{tabular}
\end{table}

\noindent
In both zero-shot directions the gain in a larger margin comes using the M-NMT model at \textit{R1}. This is the first model trained after the inclusion of the dataset generated by the \textit{dual-inference} stage. The It$\rightarrow$Ro direction improves by $+10.50$ BLEU points from a $4.72$ to $15.22$, whereas Ro$\rightarrow$It improves from a baseline score of $5.09$ to $16.31$ BLEU ($+11.22$). The contribution of the self correcting process can be seen in the subsequent rounds, i.e., the
improvements after each inference stage suggest that the generated
data are getting cleaner and cleaner. With respect to the Transformer model pivoting results shown in Table~\ref{pm-nmt2}, our approach outperformed both single pair and multilingual pivoting methods at the second round (\textit{R2}) (see the third column of Table~\ref{4direction_zst}). Compared with the better performing multilingual pivoting, our approach at the fifth round (\textit{R5}) has a $+3.34$ and $+3.02$ BLEU gain for the It$\rightarrow$Ro and Ro$\rightarrow$It directions respectively.  

\begin{table}[h!]
\centering
\caption{Results summary comparing the performance of systems trained using parallel data (i.e., two single language pair \textit{NMT} and a six direction multilingual \textit{M6-NMT} systems) against the four directions multilingual baseline (\textit{M4-NMT}) and our approach at the fifth round \textit{R5}. Best scores are bold highlighted, whereas statistically significant (p$<$0.05) results in comparison with the baseline (\textit{NMT}) are indicated with $\star$} \label{nmt_m-mnmt_r5}
\begin{tabular}{l c c c c c c r}
\hline
\textit{Direction} & \textit{NMT} & \textit{M6-NMT} & $\Delta_{\tt NMT}^{\tt M6-NMT}$ & \textit{M4-NMT} & $\Delta_{\tt NMT}^{\tt M4-NMT}$ & \textit{R5} & $\Delta_{\tt NMT}^{\tt R5}$    \\ \hline
It$\rightarrow$Ro  & 20.10        & \bf 20.13     & +0.03        & 4.72      & -15.38             & 20.11    &  +0.01    \\
Ro$\rightarrow$It  & 21.36        & 21.81         & +0.04         & 5.09     & -16.27             & \bf{22.41$^{\star}$}   & +1.05 \\ \hline
\end{tabular}
\end{table}

In addition to outperforming the pivoting mechanism, an interesting trend arises when we compare our approach with the results of the single language pair and multilingual models reported in Table~\ref{biVSm-nmt2}. The summary in Table~\ref{nmt_m-mnmt_r5} shows 
the effectiveness of
a dual-inference mechanism in allowing the model to learn from its outputs. Compared to the models trained using parallel data (i.e., \textit{NMT} and \textit{M6-NMT}), our approach (\textit{R5}) is either comparable ($+0.01$ BLEU in It$\rightarrow$Ro) or better performing ($+1.05$ BLEU in  Ro$\rightarrow$It). The trend from the \emph{train-infer-train} stages indicates that, with additional rounds, our approach can further improve the dual translations. Overall, our iterative self-learning approach showed to deliver better results than the bilingual counterparts within five rounds, where each rounds iterates for a maximum of three epochs. Indeed, the improvement from our approach is a concrete example to train models in a self-learning way, potentially benefiting language directions with a parallel data, if casted in a similar setting.

\section{Conclusions}
\label{sec:concl}
In this paper, we used a multilingual NMT model in a low-resource language pairs scenario. 
Integrating and extending the work presented in \cite{MNMTlow-resourceSurafel} and \cite{lakew2017improving}, we showed that a single multilingual system outperforms bilingual baselines while avoiding the need to train several single language pair models. \mf{In particular, we  confirmed the superiority of transformer over recurrent NMT architectures in a multilingual setting.} For enabling and improving a zero-shot translation, we showed \textit{i)} how a multilingual pivoting can be used for achieving comparable results to those of multiple bilingual models, and \textit{ii)} that
our proposed self-learning procedure boosts performance of multilingual zero-shot directions by even outperforming both pivoting and bilingual models. In future work, we plan to explore our approach across language varieties using a multilingual model.

\section*{Acknowledgements}
This work has been partially supported by the EC-funded projects ModernMT (H2020 grant agreement no. 645487) and QT21 (H2020 grant agreement no. 645452). This work was also supported by The Alan Turing Institute under the EPSRC grant EP/N510129/1 and by a donation of Azure credits by Microsoft. We gratefully acknowledge the support of NVIDIA Corporation with the donation of the Titan Xp GPU used for this research.

\bibliographystyle{fullname}
\bibliography{bibliography}

\end{document}